\documentclass[pmlr]{jmlr}

\usepackage{longtable}

\usepackage{booktabs}
\usepackage{graphicx}
\usepackage{multirow}
\usepackage{adjustbox}
\usepackage{hyperref}
\usepackage{longtable}

\usepackage{color,soul}
\usepackage[load-configurations=version-1]{siunitx} 

\makeatletter
\def\set@curr@file#1{\def\@curr@file{#1}} 
\makeatother

\theorembodyfont{\upshape}
\theoremheaderfont{\scshape}
\theorempostheader{:}
\theoremsep{\newline}

\jmlrvolume{182}
\jmlryear{2022}
\jmlrworkshop{Machine Learning for Healthcare}

\title[Density-Aware Personalized Training for Risk Prediction in Imbalanced Medical Data]{Density-Aware Personalized Training for Risk Prediction in Imbalanced Medical Data}

\author{\Name{Zepeng Huo\textsuperscript{1}}
       \Email{guangzhou92@tamu.edu}\\ 
       \Name{Xiaoning Qian\textsuperscript{2}}
       \Email{xqian@ece.tamu.edu}\\ 
       \Name{Shuai Huang\textsuperscript{3}}
       \Email{shuaih@uw.edu}\\ 
       \Name{Zhangyang Wang\textsuperscript{4}}
       \Email{atlaswang@utexas.edu}\\ 
       \Name{Bobak J. Mortazavi\textsuperscript{1}}
       \Email{bobakm@tamu.edu}\\ 
       \\
       \addr \textsuperscript{1}Computer Science \& Engineering Department, Texas A\&M University, College Station, TX, USA \\
       \addr \textsuperscript{2}Electrical \& Computer Engineering Department, Texas A\&M University, College Station, TX, USA \\
       \addr \textsuperscript{3}Industrial \& Systems Engineering Department, University of Washington, Seattle, WA, USA \\
       \addr \textsuperscript{4}Electrical \& Computer Engineering Department, University of Texas, Austin, TX, USA \\
} 

\begin{document}

\maketitle

\begin{abstract}
  Medical events of interest, such as mortality, often happen at a low rate in electronic medical records, as most admitted patients survive. Training models with this imbalance rate (class density discrepancy) may lead to suboptimal prediction. Traditionally this problem is addressed through ad-hoc methods such as resampling or reweighting but performance in many cases is still limited. We propose a framework for training models for this imbalance issue: 1) we first decouple the feature extraction and classification process, adjusting training batches separately for each component to mitigate bias caused by class density discrepancy; 2) we train the network with both a density-aware loss and a learnable cost matrix for misclassifications. We demonstrate our model's improved performance in real-world medical datasets (TOPCAT and MIMIC-III) to show improved AUC-ROC, AUC-PRC, Brier Skill Score compared with the baselines in the domain.
\end{abstract}

\section{Introduction}

\noindent Machine learning-based medical risk prediction models continue to grow in popularity \cite{zhang2018patient2vec, rajkomar2018scalable, miotto2016deep}. However, the performance of these models is often biased in evaluation by commonly reported metrics (such as area under the curve of the receiver operating characteristic: AUC-ROC), often reporting overly-optimistic findings as a result of the imbalance between those that observe \textit{medical adverse events} and those that do not \cite{swets1979roc, lobo2008auc, cook2007use, huang2021performance}. 
The adverse event of interest is often in the minority class \cite{li2010learning}. For example, in mortality prediction, patients with higher risk represent a smaller fraction in the cohort compared to most of the people who survive. Naively applying machine learning models may render dissatisfaction: the outcome of interest can be extremely costly, either through unnecessary medical intervention (type 1 error) or misdiagnosis (type 2 error). Furthermore, it is important not only to rank expired patients higher than survived patients w.r.t. probability output (e.g. AUC-ROC), but also the probability output is more calibrated \cite{park2020califorest}.

While methods to tackle this imbalance issue via resampling or reweighting methods constitute a popular approach \cite{wang2020mimic, babar2016mlp}, their applications in the context of medical data are often heuristic (or case by case) in nature. First, these techniques may give readjusted importance to the smaller class, but the weighting ratio remains ad-hoc from dataset to dataset, therefore, manual tuning might not be ideal. Second, apart from inter-class density discrepancy, one unique aspect of medical data is that even in the same risk group (same label), the patients may have different underlying comorbidities or risk factor characteristics that arrive at potentially high risk for various reasons, rendering intra-class heterogeneity~\cite{huo2019sparse}. This heterogeneity requires models to have a personalized training regime to distinguish the nuanced differences \cite{huo2021sparse}, to address the imbalance in a standardized/automated fashion. Rather than treating imbalanced densities as a problem, exploiting this information in training may enhance performance \cite{ali2013classification}.

We propose a framework to address class imbalance density and make use of this imbalance to render density-aware training for improved risk prediction performance. First, we decouple the training of representation learning and classification. Traditionally, representation learning and classification are trained jointly \cite{kang2019decoupling}, but by decoupling, class-specific features are extracted and class-specific predictions made, removing a source of bias for the learned classifier \cite{zhou2020bbn}. 
Second, the density differences are important to learn, not eliminate, when modeling. Patients with lower risk (majority) are often lower risk because they do not contain any of the common risk factors (e.g. lack of hypertension, diabetes, prior myocardial infarction), and hence, form a dense cluster. However, patients with higher risk (minority) may arrive at this high risk from different factors (e.g. renal failure versus respiratory distress), thus being scattered in the data space \cite{huo2021sparse}. Our approach is density-aware, by avoiding re-sampling or re-weighing pre-processing steps, and the decoupling approach improves risk prediction performance. We demonstrate this approach in two different medical data scenarios: a randomized clinical trial dataset and an electronic health record (observational) dataset. We show that our method can achieve high predictive performance in these imbalanced medical datasets (imbalance ratio can range from 7 $\sim$ 10) and perhaps surprisingly it can also achieve superior calibration than the baselines without an extra set of calibration data.

\vspace{3mm}
\noindent \textbf{Generalizable Insights about Machine Learning in the Context of Healthcare}

\noindent
As sophisticated models for increasingly large medical datasets are developed and promoted, evaluation of the predicted outcomes, through the use of an appropriate set of metrics is necessary. Medical data often contains low event rates for the major adverse events of interest. The primary measures of their performance, either threshold specific-based classification techniques, which may not properly account for the different costs of Type I and Type II errors, or the ability of the model to discriminate those at risk and those not, through the AUC-ROC, become overconfident in telling clinicians using the model who is not at risk. By leveraging the imbalance in the classes modeled, we are able to more accurately estimate those at risk, by more accurately identifying driving risk factors in the groups independently. As a result, this allows us to more concretely evaluate why they individuals are at risk (rather than simply being not at low-risk), and provide for better model calibration for medical decision making - through probabilistic interpretation of an occurrence in a frequentist perspective. 


\section{Related Work}
Supervised learning methods on imbalance dataset tasks often re-balance data via re-sampling, such as oversampling \cite{pouyanfar2018dynamic}, undersampling \cite{he2009learning} classes. Others use synthetic samples to account for imbalance, where new samples are generated from perturbations of old samples \cite{chawla2002smote, zhang2018mixup}. Another common approach is via re-weighting, which re-assigns training weights for each class based on criteria such as number of instances of each class \cite{huang2019deep}, effective numbers \cite{cui2019class} or the distance between loss \cite{cao2019learning}. However it is not clear the clinical utility of these methods since they were developed in non-medical datasets.

Medical models often focus on risk of adverse event estimation, which intrinsically carries data imbalance. Re-sampling has been widely applied \cite{chawla2002smote, bhattacharya2017icu}. Cost-sensitive training is also applied, for example on Intensive Care Unit (ICU) data \cite{rahman2013addressing}. Hybrid approaches, which combine re-sampling and cost-sensitive training have also been applied \cite{li2010learning}. These methods all are based upon the ad hoc tuning, weighting, or re-sampling to address imbalance, but do not learn from the imbalance information itself.

Furthermore, the imbalance issue in medical dataset not only affects the prediction, but also calibration \cite{park2020califorest}. The currently used metrics in medical modeling are usually not geared towards calibration and the metrics most widely used, such as AUC-ROC is susceptible to imbalance ratio  \cite{huang2021performance}. The modern-day neural networks have achieved astonishing accuracy but studies have shown most methods are getting less and less calibrated \cite{guo2017calibration}. Therefore we will demonstrate in our model the calibration is an extra contribution on top of handling imbalance prediction


\section{Methods}
\begin{figure*}[ht!]
\centering
\includegraphics[width=\textwidth]{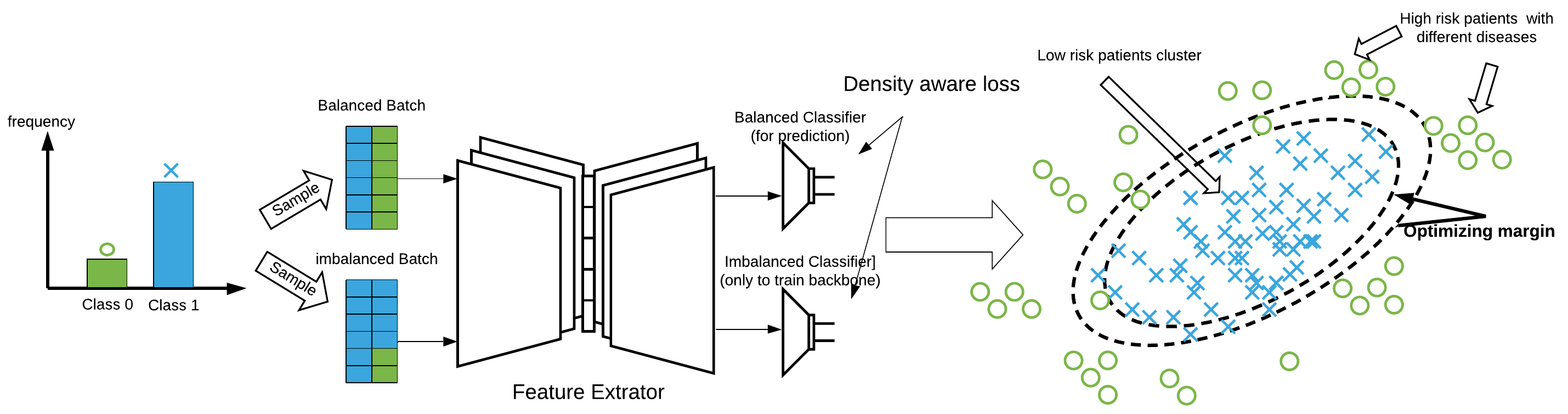}
\caption{Overall pipeline. The original medical dataset is sampled with two distributions, one balanced and one imbalanced (default) batch. Both of the batches will go through the backbone but the classifiers will utilize different batches to optimize the density-aware loss, rendering personalized decision boundaries for different classes.}
\label{fig:pipline}
\end{figure*}

In this section, we introduce our framework. The approach first separates the training data to different risk groups. Then, it uses a density-aware loss function to take into account the data density difference between majority class and minority class. Finally, it uses a learnable cost matrix to personalize misclassification. We stress that our framework is a training regime that can apply to different backbones (e.g. different neural network architectures) and we will later show in experiments this framework being used on real-world tabular as well as time-series medical data.
The overall pipeline is shown in Fig \ref{fig:pipline}.

\subsection{Decoupling training for imbalance classes}
In neural networks, we can coarsely define the last layer (or last few layers) \citep{zhou2020bbn} as the output classifier, since the output is used to determine the class of one specific instance. The previous layers of the network architecture can be deemed as the feature extractors or backbone. Traditionally these two parts are trained jointly and the distinctions between them are ill-defined \citep{kang2019decoupling}. However, \cite{zhou2020bbn} showed that the classifier portion of the network is more susceptible to data imbalance, whereas the feature extractor is not, during training. Thus, we decouple training of the feature extractor and classifier. 

Formally, let $\mathcal{I}= \{I_i\}$ be a set of inputs, and $\mathcal{Y}=\{y_i\}$ be the set of corresponding labels. For a typical objective function, we write:
\begin{equation}
    \mathcal{L} = \frac{1}{N}\sum_{c=1}^{|C|}\sum_{i=1}^{|N_c|}l(h(I_{ci}), y_{ci}),
\end{equation}
where $l(\cdot)$ is the loss function and $h(\cdot)$ is our model. $|C|$ signifies the number of total classes and $|N_c|$ the number of instances in one specific class. In an imbalance setting, the larger class with more training instances $|N_c|$ will dominate the loss and thus make the model biased. A naive way to tackle this issue is to adjust the sampling rate for the smaller class. For example, a \textit{class-balanced sampling} (CBS) is proposed \citep{huang2016learning}, where the instances from each class are sampled with equal probability so that the big class would not dominate the loss calculation and hence the density discrepancy will have much less effect. 
However, the CBS strategy will likely induce the ill-fitting problem because either the big class is under-sampled, inducing loss of information or small class is over-sampled, inducing over-fitting \citep{yang2020rethinking}. We propose that for a well-trained neural network, a set of abundant and diverse training instances is required, so that the model can generalize well in the testing set. 
A method is required to make use of the rich information in the big class but to ensure the smaller class is well represented as well.

Inspired by \cite{zhang2019balance}, we are proposing a solution to use both \textit{class-balanced sampling} and \textit{regular random sampling}, where the first would sample each instance to make sure each class has equal probability and the latter one samples each instance with equal probability. The function can be defined:
\begin{equation}
    p_j = \frac{n_j^q}{\sum_{c=1}^{|C|} n_c^q},
\end{equation}
where $p_j$ indicates the probability of sampling a data point from class $j$ and the range for $q \in [0, 1]$ and $|C|$ is the number of classes. The \textit{regular random sampling} entails the $q = 1$, meaning the probability will be proportional to the cardinality of the class $j$. The \textit{class-balanced sampling} would entail $q = 0$ which means $p_j = 1/ |C|$, and therefore each class is balanced. These two sampling strategies will generate two sets of batches, with each class's density built in, and we will train the feature extractor with both batches while the classifier will only train on the corresponding batch. In this way the rich information of big class will be preserved and at the same time the balanced classifier, which is eventually used for prediction during inference, will not be biased towards one class. 

\subsubsection{Density-aware outlier detection loss}
To further make use of the inherent density information among the classes, we will introduce the density-aware training. There have been many cost-sensitive methods proposed to address the imbalance issue. One of the most popular ones is the focal loss \citep{lin2017focal}. This method focuses on the `difficult' examples, which means the predicted probability of the example is far away from the true label. Based on previous discussion, we can treat the low risk patient as in-distribution data and high risk patients (with different underlying factors) as out-of-distribution data, and use the outlier detection technique to optimize the boundary \citep{huo2020uncertainty}.

By following this direction we propose a hinge loss based objective function. Hinge loss itself is less susceptible to density discrepancy among classes because it aims to optimize around support vectors, thus focusing the `difficult' examples which are close to the decision boundary. However the traditional `max-margin' training using the hinge loss did not take into account the class-wise density, which renders a non-personalized training. Our proposed personalized training is through a density-aware margin optimization \citep{cao2019learning}. This Density-Aware Hinge (DAH) loss can be written as follows:
\begin{equation}
\begin{aligned}
    \mathcal{L}_{DAH} &=\frac{1}{N} \sum_c^{|C|} \max(\max_{j\neq c}\{z_j\} -z_c + \Delta_c, 0) \\
    \text{where} \ & \Delta_c = \frac{\text{K}}{|N_c|^{1/4}} \text{, for}\ c \in \{1, ..., |C|\}, 
\end{aligned}
\end{equation}
where $\mathcal{L}_{DAH}$ is the density-aware hinge loss, $z_j$ is the model $j$-th element in the output vector, indicating the probability of this instance predicted to be $j$-th class, and $z_c$ is the predicted output probability of the true class $c$. The form follows the traditional hinge loss, except the density-aware component $\Delta_c$. The parameter $\text{K}$ is a hyper-parameter, and $|N_c|$ is number of examples in class $c$. In \cite{cao2019learning}, the exponential in $|N_c|^{1/4}$ is derived by the trade-off of optimizing all the margins between classes, so that the imbalanced test error can be smaller than a generalization error bound \citep{wei2019improved}. That is, $\gamma_j \propto |N_c|^{1/4}$, where $\gamma_c$ is the margin in the hyper-plane for class $c$. Therefore we follow this tuning. The hyper-parameter $\text{K}$ is usually tuned by normalizing the last hidden activation and last fully-connected layer's weight vectors' $\ell_2$ norm to 1, as noted in \cite{wang2018additive}.

In practice, the hinge loss may pose difficulty for optimization due to its non-smoothness \citep{luo2021learning}. First we derive the softmax from the original form and thus a relaxed form of hinge loss for smoothness is adopted to simulate the cross-entropy form:
\begin{equation}
\begin{aligned}
    \mathcal{L}_{DAH} &=\frac{1}{N}\sum_c^{|C|} -\log \sigma(z_c),  \\
     \text{where }  & \sigma(z_c)= \frac{\text{exp}(z_c-\Delta_c)}{\text{exp}(z_c-\Delta_c)+\sum_{j \neq c}\text{exp}(z_j)},  \text{and } \Delta_c = \frac{\text{K}}{|N_c|^{1/4}} \text{, for}\ c \in \{1,...,|C|\}
\end{aligned}
\end{equation}
The `max-margin' form is relaxed to a softmax function in the cross-entropy-like optimization. While some previous work \citep{liu2016large, wang2018additive} adopted similar ideas, our proposed personalized margin $\Delta_c$ can make use of the information in density discrepancy itself for training.

\begin{table}[t] 
\caption{A typical cost matrix where the diagonal has zero cost, and $C_{FN}$, $C_{FP}$ represents false negative cost and false positive cost, respectively.}\vspace{-0.05in}
\begin{center}
\begin{normalsize}
\begin{tabular}{c|c|c}
\hline
\small Cost Matrix   & Predicted as Positive &   Predicted as Negative \\
\hline
\small True Positive & 0 & $C_{FN}$ \\
\small True Negative & $C_{FP}$ & 0\\ 
\hline
\end{tabular}
\label{cost_M}
\end{normalsize}
\end{center}
\vskip -0.1in
\end{table}

\subsubsection{Trainable cost matrix}
For personalized training, we propose to equip the density-aware loss with a trainable cost matrix. Traditionally the cost of training has been set static throughout the whole training process (e.g. false positive cost and false negative cost in binary classification). The default cost matrix can be seen as table \ref{cost_M}, where the $C_{FN}$, $C_{FP}$ were traditionally set to 1 (Note that the cost matrix here can only be applied to binary prediction). However this implies that the two types of cost are equal throughout the whole training \citep{roychoudhury2017cost}. But as we discussed before, the big class and small class would make the model more biased towards one versus the other due to the density disparity. But we want to use some mechanism to rebalance the training so that the model would be less biased. Thus instead of treating the costs as a prior knowledge, we make them as trainable parameters along with the model as well \citep[page.~66]{fernandez2018learning}. In this way, the model will dynamically learn the cost to minimize the loss function.
For an input and target pair $(x, y)$, where the output of the model is $z=h(x)$, the loss function with incorporation of two costs under binary classification is proposed:
\begin{equation}
\begin{aligned}
    \mathcal{L}((x,y);h(\cdot)) = &-y\log\sigma(C_{FN}z_{\max}) - (1 -y)\log (1-\sigma(C_{FP}z_{\max})) \\
    \text{subject to } & C_{FN} > 0, C_{FP} > 0, C_{FN} > \theta C_{FP}
\end{aligned}
\label{trainable_cost}
\end{equation}
The $z_{\max}$ indicates the largest logit along the output vector. The constraints above ensure that the two types of misclassification cost will always be positive \citep{roychoudhury2017cost} and due to the minority class is the prediction of interest (such as higher risk patients), we penalize more in the event of false negatives verse false positives. Here, $\theta$ can be tuned as a hyper-parameter.

In practice, when applying stochastic gradient descent (SGD), the parameters can only be updated without constraints. Here we relax the constrained problem as an unconstrained one, we thus rewrite:
\begin{equation}
    C_{FN} = \theta C_{FP} + \mathcal{D},
\end{equation}
where $\mathcal{D}$ is a regularization term. Therefore we will only need to make sure $C_{FP} > 0$ during training. We propose to  minimize the objective loss function in terms of $\log C_{FP}$ instead of $C_{FP}$:
\begin{equation}
    \frac{\partial \mathcal{L}((x,y);h(\cdot))}{\partial \log C_{FP}} =  C_{FP}\frac{\partial \mathcal{L}((x,y);h(\cdot))}{\partial C_{FP}}, 
\end{equation}
where the loss function can take the form as we defined above for  density-aware training. Note that there are generally two ways to handle the constraints for optimization: reparameterization to an unconstrained minimization problem
or projected gradient (PG) \citep{amid2020reparameterizing}. PG is to perform unconstrained gradient updates, then project back onto the feasible space after each update. PG directly solves the convex optimization problem, but the intermediate iterates can sometimes lead to a possibly less stable or too aggressive trajectory \citep{raskutti2014information}. Ours is similar to reparameterization where numerical stability is more warranted in this regard.


\begin{table}[t] 
\centering
\caption{Summary of the datasets and the tasks}

\begin{tabular}{c|c|c|c|c}
\hline
Dataset & task   & \#instances & \#features & IR  \\
\hline
\multirow{2}{*}{TOPCAT} & Mortality & \multirow{2}{*}{1,767} & \multirow{2}{*}{86} & 7.92\\
& hospitalization & & & 1.71\\
\hline
\multirow{2}{*}{MIMIC-III} & Mortality & \multirow{2}{*}{21,139}
 &\multirow{2}{*}{34} & 7.57\\
& Phenotyping & & & 10.32\\
\hline
\end{tabular}
\label{dataset}
\end{table}

\section{Experiments}
\subsection{Datasets}
In our experiment, we test our proposed model on two real-world medical datasets which include inherent imbalance issues and heterogeneous patients representations.

1) The first dataset is TOPCAT (Treatment of Preserved Cardiac Function Heart Failure with an Aldosterone Antagonist). TOPCAT is a multi-center, international, randomized, double-blind, placebo controlled trial sponsored by the U.S. National Heart, Lung, and Blood Institute \cite{bertram2014spironolactone}. TOPCAT collects patients from the United States, Canada, Brazil, Argentina, Russia, and Georgia between 2006 and 2013. The outcomes of interest were all-cause mortality and heart failure hospitalization through 3 years of follow-up. The data includes demographic and clinical data available from patients in addition to laboratory data, electrocardiography data, Kansas City Cardiomyopathy Questionnaire (KCCQ) scores (physical limitation score, symptom stability score, symptom frequency score, symptom burden score, total symptom score, self-efficacy score, quality of life score, social limitation score, overall summary score, and clinical summary score). The details of the variables are listed in supplementary Table \ref{tab:TOPCAT_variables}. 

2) The second dataset is MIMIC-III (Medical Information Mart for Intensive Care). MIMIC-III is one of the largest clinical datasets that has been made publicly available \cite{johnson2016mimic}. It contains multivariate time-series data from over 40,000 intensive care unit (ICU) stays. The types of data range from static demographics such as gender and age to rapidly changing measurements such as heart rate and arterial blood pressure. The heterogeneity is one of the major challenges when analyzing this dataset, due to the diverse patient health conditions, rapidly changing hazard ratio as well as the corresponding treatments. We focus on using only the first 48 hours after ICU admission for the prediction of patient mortality and phenotyping. The intuition is that for early risk prognosis and phenotyping, the precaution procedure can be undertaken since the average ICU stay can be up to 100 to 200 hours \cite{johnson2016mimic}. We adopted the same data pre-processing steps as in a set of benchmark models in MIMIC-III  \cite{harutyunyan2019multitask}(i.e. imputation, normalization, data masking, etc), where we used the same 17 clinical measurements and their derivations to construct in total 34 time-series features.

A summary of the datasets with the imbalance ratio (IR) is shown in Table \ref{dataset}. We test our model on three binary classifications (2 from TOPCAT, 1 from MIMIC-III) tasks and a multi-class classification task (MIMIC-III) to demonstrate our model can work under a variety of scenarios. Note that only the phenotyping task in MIMIC-III is a multi-class multi-label scenario, so the imbalance ratio is calculated between the largest class and smallest class. The listing of phenotype labels used is in supplementary Table \ref{MIMIC_pheno}, along with their medical type to indicate this is a heterogeneous set of labels that have many underlying driving factors.  
The model is not trained on a learnable cost matrix for phenotyping since the false positive and false negative is for binary classification, therefore, we solely rely on decoupling training and density aware loss.

\subsection{Experimental Setup}
For each of the datasets and tasks, we selected strong baselines from existing benchmarks.

1) Baselines for TOPCAT:
\begin{itemize}
\setlength\itemsep{-0.5em}
    \item RF \cite{angraal2020machine}: a Random Forest based method which is originally tested on TOPCAT dataset
    \item U-RF \cite{arafat2017cluster}: a balanced Random Forest that randomly under-samples each boostrap sample to balance training
    \item R-MLP \cite{babar2016mlp}: a Multi-layer Perceptron (MLP) model that uses reweighting in the training 
\end{itemize}
For all the baseline with resampling or reweighting, we train the network on 80\% of the data and tune the hyper-parameters including the weighting ratio in 10\% of the data, and test on the rest 10\%. For our model to have a strong neural work backbone, we construct a multi-layer Perceptron model as our backbone. The construction of the backbone is similar to R-MLP \cite{babar2016mlp}, training the neural network with 200 epochs, with learning as 0.001 and batch size as 64. More specifically this is a 4-layer fully connected NN, the first input layer is the same as number of features and each hidden layer has 28 neurons with one residual skip connection block and output layer has 2 neuron which is later measured on cross-entropy loss. We have our model train with our proposed decoupling and density aware loss, whereas R-MLP has under-sampling as their technique with some stochastic measures.

2) Baselines for MIMIC-III:
\begin{itemize}
\setlength\itemsep{-0.5em}
    \item GRU-D \cite{che2018recurrent}: a Gated Recurrent Unit (GRU) based method where the model has a trainable decay component
    \item bi-LSTM \cite{harutyunyan2019multitask}: a bi-directional Long Short-term Memory (LSTM) based method with channel-wise feature fusion
    \item flexEHR \cite{deasy2019impact}: a GRU based method that uses word embedding technique to extract features.
    \item GRU-U \cite{wang2020mimic}: a GRU based method that utilizes both trainable decay and undersampling technique for imbalance handling
    \item c-LSTM \cite{harutyunyan2019multitask}: a channel-wise LSTM that process each variable independently in the first layer then fuse them in the second layer
    \item Deep Supervision \cite{lipton2015learning}: an RNN based model that uses target replication for the supervision of LSTM in each time stamp, and with changing loss function the model needed to predict replicated target variables along with outcome
\end{itemize}
For the MIMIC-III dataset, we follow the same 80/10/10 splits. And we construct our backbone same as flexEHR \cite{deasy2019impact} which is a GRU based method. We trained the models with 50 epochs with an early stopping threshold of 5 epochs with no increase in AUC-ROC on the validation set. The batch size is 128 and Adam optimizer is used with learning rate 0.001.

In addition to the traditional way of measuring probabilistic output of the medical models, i.e. area under the receiver operating curve (AUC-ROC), we argue that we need to incorporate the metrics that can represent the difficulties induced by imbalanced class densities. First AUC-ROC only measures the true positives (TP) and false positive (FP) relationship, which can present an overly optimistic view of an algorithm's performance if there is large skew in the class distribution \cite{davis2006relationship}. 
On the other hand, area under precision-recall curve (AUC-PRC) can provide a more reliable interpretation under imbalance, due to the fact that they evaluate the fraction of true positives among positive predictions \cite{saito2015precision}, and the precision-recall relationship will change when the test set's imbalance ratio changes, thus providing more sensitive evaluation \cite{davis2006relationship}. 
Furthermore, in a medical model, the conventional way of measuring the model is through Brier score \cite{brier1950verification}, which takes into consideration the calibration of the model. However, the Brier score is also susceptible to imbalance ratio \cite{fernandez2018learning}. We propose to use Brier Skill Score (BSS) \cite{fernandez2018learning}, where the model takes the calculated Brier score and compare it to a reference point, i.e. a scaled Brier score by its maximum score under a non-informative model \cite{steyerberg2010assessing}, to show the improvement:
\begin{equation}
    BSS = 1 - \frac{BS}{BS_{\text{max}}}
\end{equation}
We chose the reference $BS_{\text{max}}$ to the the prevalence predictor to output the probability based on the imbalance ratio, i.e. $BS_{\textrm{max}}=\frac{1}{N}\sum_{t=1}^N(f_t-o_t)^2$ and prediction $f_t$ is replaced by the event rate and $o_t$ is the outcome label of interest \cite{center2005brier}.

\begin{figure}[t!]
\centering
\begin{minipage}{.7\textwidth}
\centering
  \includegraphics[width=\textwidth]{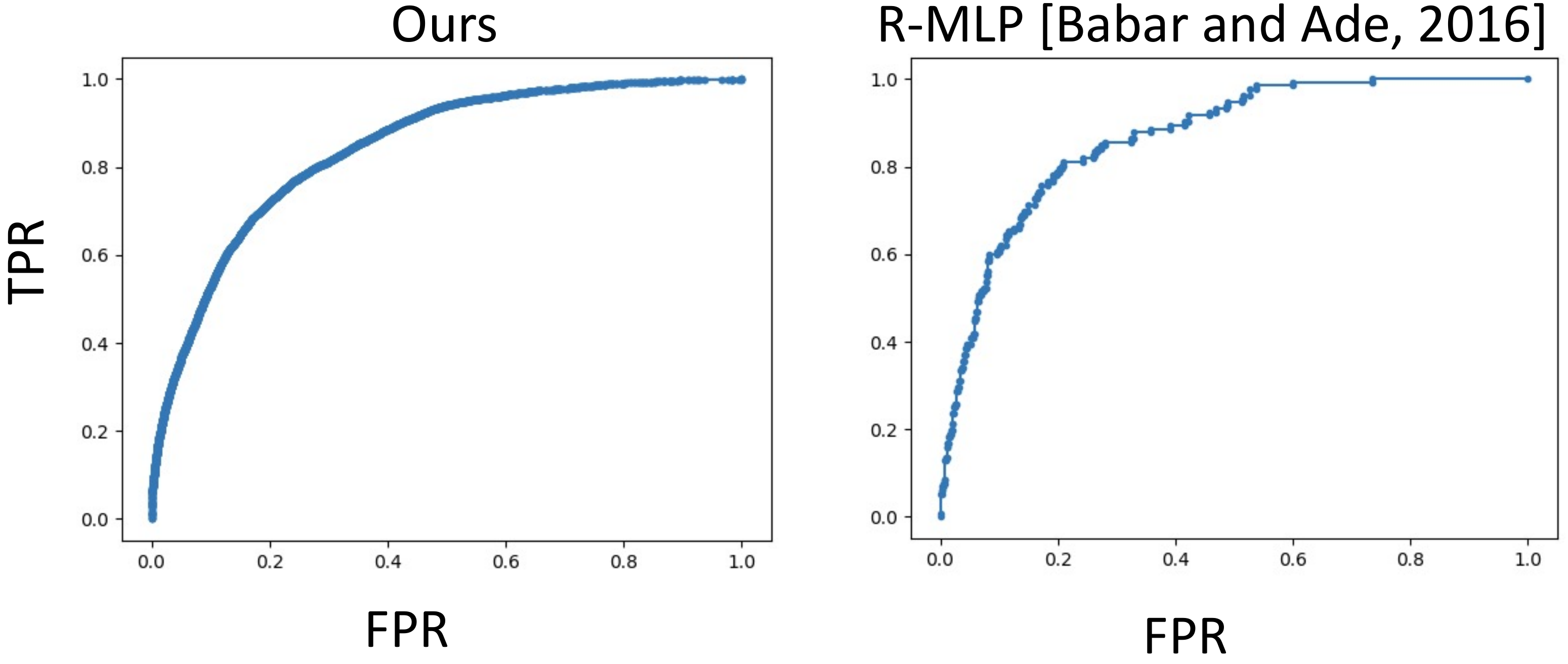}\vspace*{-3mm}
  \caption{AUC-ROC comparison}
  \label{fig:auroc}
\end{minipage}
\hfill
\begin{minipage}{.7\textwidth}
\centering
  \vspace*{6mm}
  \includegraphics[width=\textwidth]{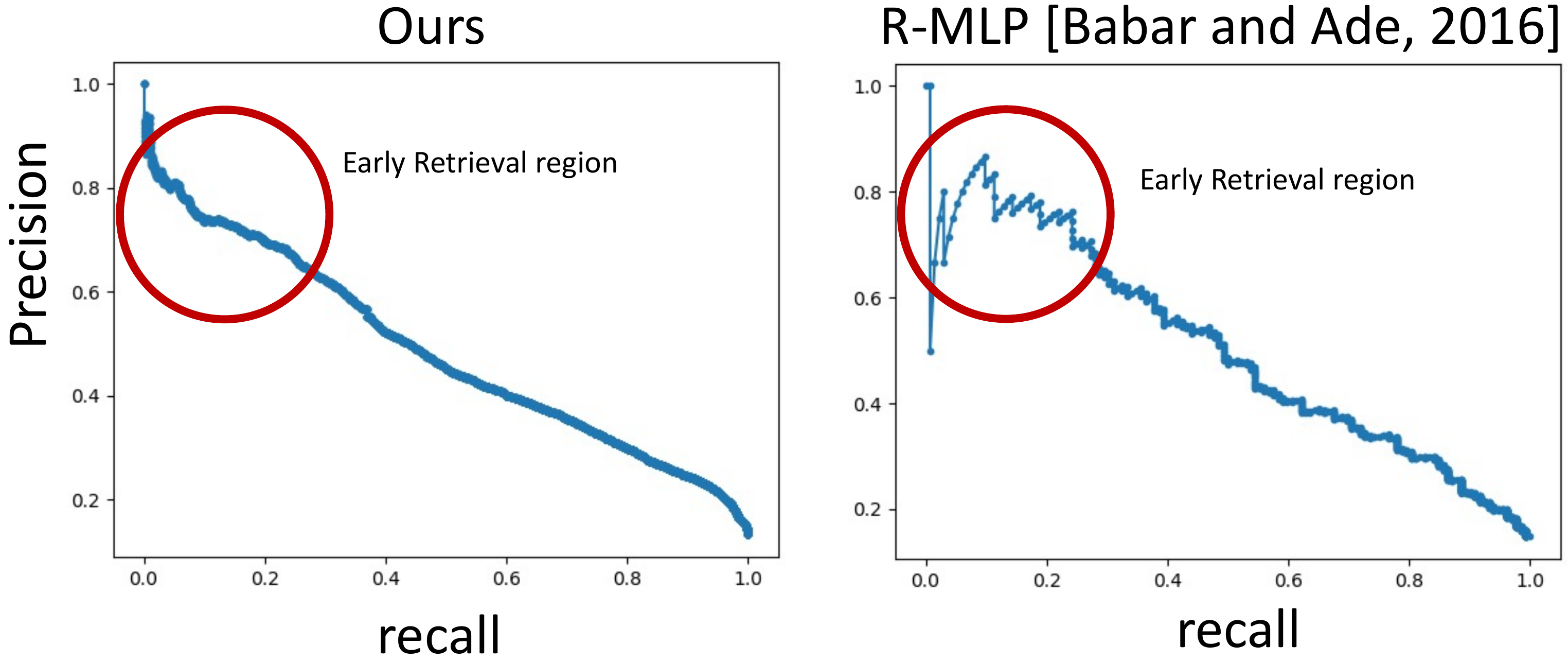}\vspace*{-3mm}
  \caption{AUC-PRC comparison}
  \label{fig:auprc}
\end{minipage}%
\end{figure}

\begin{table*}[t!]
\small
\centering
\caption{Results for TOPCAT dataset}
\label{tab:TOCAT_result}
\begin{tabular}{|l|l|l|l|l|}
\hline
Task & Methods  & AUC-ROC & AUC-PRC & BSS  \\ \hline
\multirow{5}{*}{Mortality} &
RF \cite{angraal2020machine}    & 0.723 $\pm$ 0.003  &  0.512 $\pm$ 0.001  &     -0.357 $\pm$ 0.002    \\
& U-RF \cite{arafat2017cluster}    & 0.752 $\pm$ 0.002  &  0.532 $\pm$ 0.002   &     -0.103  $\pm$ 0.003     \\
& R-MLP \cite{babar2016mlp}     & 0.736 $\pm$ 0.001  &   0.523 $\pm$ 0.005  &    -0.067 $\pm$ 0.003       \\
\cline{2-5}
& Ours   & \textbf{0.794 $\pm$ 0.002}  &  \textbf{0.583 $\pm$ 0.002}   &      \textbf{0.166 $\pm$ 0.003}     \\
\hline
\hline
\multirow{5}{*}{Hospitalization}&
RF \cite{angraal2020machine}     & 0.763 $\pm$ 0.005  & 0.657  $\pm$ 0.006   & -0.008    $\pm$ 0.0004    \\
& U-RF \cite{arafat2017cluster}     &  0.771 $\pm$ 0.005  &  0.674 $\pm$ 0.006  & -0.005  $\pm$ 0.003   \\
& R-MLP \cite{babar2016mlp}  & \textbf{0.789  $\pm$ 0.003} &  0.661 $\pm$ 0.005 &     -0.012 $\pm$ 0.001          \\
\cline{2-5}
& Ours  &  0.788 $\pm$ 0.007  &  \textbf{0.711 $\pm$ 0.003}  &   \textbf{0.132  $\pm$ 0.002}      \\
\hline
\end{tabular}
\end{table*}

\vspace{-3mm}

\begin{table*}[t!]
\small
\centering
\caption{Results for MIMIC-III dataset}
\label{tab:MIMIC_result}
\begin{tabular}{|l|l|l|l|l|}
\hline
Task & Methods & AUC-ROC & AUC-PRC & BSS  \\ \hline
\multirow{6}{*}{Mortality} &
GRU-D \cite{che2018recurrent}      & 0.852 $\pm$ 0.002   &  -   &     -     \\
& bi-LSTM \cite{harutyunyan2019multitask}     & 0.862 $\pm$ 0.004  & 0.515 $\pm$ 0.001  &  -0.801  $\pm$ 0.002       \\
& flexEHR \cite{deasy2019impact}    & 0.878 $\pm$ 0.004  &  0.513 $\pm$ 0.002  &  -1.105  $\pm$ 0.003     \\
& GRU-U \cite{wang2020mimic}     & 0.876 $\pm$ 0.006 &  0.532 $\pm$ 0.002   &  -        \\
\cline{2-5}
& Ours    & \textbf{0.892 $\pm$ 0.001}  & \textbf{0.586$\pm$ 0.004}   &   \textbf{0.240 $\pm$ 0.003}      \\
\hline 
\end{tabular}
\begin{tabular}{|l|l|l|l|}
\hline
Task & Methods & Macro AUC-ROC & Micro AUC-ROC   \\ \hline
\multirow{3}{*}{Phenotyping}&
c-LSTM \cite{harutyunyan2019multitask}      &   0.708 $\pm$ 0.0023  & 0.725 $\pm$ 0.0053  \\
 &  bi-LSTM \cite{harutyunyan2019multitask}     & 0.770 $\pm$ 0.0081  &    0.791 $\pm$ 0.0048            \\
(Multi-class, & flexEHR \cite{deasy2019impact}     & 0.755 $\pm$ 0.0052  &  0.814 $\pm$ 0.0071              \\
\multirow{1}{*}{Multi-label)} & Deep Supervision \cite{lipton2015learning}     &  0.679 $\pm$ 0.0074 &     0.713 $\pm$ 0.0061          \\
\cline{2-4}
& Ours   & \textbf{0.771 $\pm$ 0.0061}  & \textbf{0.821 $\pm$ 0.0049}              \\
\hline
\end{tabular}
\end{table*}

\subsection{Results}
First, for the TOPCAT dataset in Table \ref{tab:TOCAT_result}, we have compared our model with the baselines and we repeated the experiments in a 5-fold cross-validation scenario and compute the 95\% confidence intervals. In the mortality prediction task, we are performing better on three metrics, especially in the imbalance oriented metric, AUC-PRC and BSS. The margin improved on AUC-PRC is obvious, showing the model is sensitive on finding a balance between precision and recall, both of which measure the performance of the class of interest (minority class where the patients eventually expired). Furthermore, the baselines model all have negative BSS scores, showing that in this mortality prediction scenario, the imbalance can pose a big challenge for a model to calibrate. In fact negative BSS is not uncommon in existing work \cite{weigel2007discrete, leadbetter2022assessing}, showing that many modern-day models can have high predictive power but are poor at calibration, as noted in \cite{guo2017calibration}. We will later show that by comparing against with some post hoc calibration technique on the baselines, our model can still stand out on both prediction and calibration. Next for hospitalization prediction, our model also outperforms the baselines in two of the key metrics. The AUC-ROC is second to the best, after the same model backbone trained on resampling. We suspect this is due to the fact that the IR score is lower in this task, rendering less focus on difficulty induced by imbalance and resampling is designed to handle the example-wise difficulty. However as we discussed before, the AUC-ROC is not sensitive to class distribution so the majority class's performance can lead to the model having an overly optimistic evaluation. We compared the AUC-ROC plot and AUC-PRC plot of our model and the R-MLP baseline in Figure \ref{fig:auroc} and \ref{fig:auprc}. As can be seen, the AUC-ROC plots of the two models are similar, however, the AUC-PRC plot shows on the upper region of the curve the baseline is performing rather unstably but our model gives a more smooth curve. In \cite{saito2015precision}, this region is defined as \textit{early retrieval} region, where is usually used to measure in information retrieval application for when results of interest account for a small portion of all the corpus \cite{hilden1991area, truchon2007evaluating} (what is the model precision when recall rate is low). We can conclude our model has better performance on AUC-PRC is due to the better part on \textit{early retrieval} where it can better handle the imbalance for the class of interest.

For the MIMIC-III dataset in Table \ref{tab:MIMIC_result}, we also compared our model with the baselines (Note some metrics are empty due to the original model did not report those metrics and there is no publicly available code to replicate the results). First, for mortality prediction, our model is again better across the board among all the three metrics, especially on BSS evaluation on the model. It is perhaps surprising that traditional medical models were rarely optimized w.r.t. calibration, which however is an important metric for medicine \cite{van2019calibration}.
For the phenotyping task, due to multi-class, multi-label scenario, the previously existing methods did not adopt AUC-PRC (precision, recall are used mostly in binary classification) or BSS (Brier score is used in a 1-0 probabilistic output model for calibration purposes.) So we instead use macro AUC-ROC and micro AUC-ROC for model performance comparison. From the table, we can see our model is better than all the baselines on these two metrics, showing the multi-class multi-label imbalance scenario can also be handled by our framework.

\begin{table*}[t!]
\small
\centering
\caption{Calibration study for TOPCAT dataset}
\label{tab:TOCAT_calibration}
\begin{tabular}{|l|l|l|l|}
\hline
Methods  & AUC-ROC & AUC-PRC & BSS  \\ \hline
 RF \cite{angraal2020machine}    & 0.723 $\pm$ 0.003  &  0.512 $\pm$ 0.001  &     -0.357 $\pm$ 0.002    \\
\multicolumn{1}{|r|}{w/ Califorest \cite{park2020califorest}}    & 0.734 $\pm$ 0.003  &  0.498 $\pm$ 0.001  &     -0.052 $\pm$ 0.002    \\
\cline{1-4}
 R-MLP \cite{babar2016mlp}     & 0.736 $\pm$ 0.001  &   0.523 $\pm$ 0.005  &    -0.067 $\pm$ 0.003       \\
\multicolumn{1}{|r|}{w/ Temperature scaling \cite{guo2017calibration}}    & 0.744 $\pm$ 0.003  &  0.518 $\pm$ 0.001  &     0.102 $\pm$ 0.002    \\
\cline{1-4}
 Ours   & \textbf{0.794 $\pm$ 0.002}  &  \textbf{0.583 $\pm$ 0.002}   &      \textbf{0.166 $\pm$ 0.003}     \\
\hline
\end{tabular}
\end{table*}

In \cite{guo2017calibration}, the authors argued the confidence calibration, being the problem of predicting probability estimates representative of the true correctness likelihood, is important for classification models in many applications, but modern neural networks are becoming increasingly lacking in this respect. They proposed a \textit{temperature scaling} method to calibrate the model which is a variant of Platt scaling \cite{platt1999probabilistic}. The method is to use sigmoid function as the transformation for model's output into proper posterior probability. Our method of distribution-aware loss resembles the temperature scaling in that we have a component in the softmax to `soften' the probability similarly to the temperature variable. Furthermore the component is label distribution aware, making it particularly suitable for calibration in an imbalanced setting. We aim to compare against this method. Furthermore, the tree-based methods all performed badly especially in terms of calibration. We will use a calibration specifically designed for tree models, i.e. Califorest \cite{park2020califorest}, where the authors used Out-of-Bag (OOB) samples as the calibration set and each individual prediction is used to calculate the weights for the samples. We will use these two methods as the post hoc calibration method (i.e. the model is calibrated after finished training, this will theoretically give them more advantage since our model does not use any post hoc calibration) for the tree-based model and neural networks to study if the BSS metric for them can be increased to positive. We did our experiment on TOPCAT dataset and the task is hospital mortality prediction. The results are shown in Table \ref{tab:TOCAT_calibration}. As we can see we have equipped the tree model, i.e. Random Forest, the Califorest and R-MLP the Temperature scaling to calibrate after the training, which renders improvement on BSS for both of the cases. The RF has not been able to push BSS to positive but the improvement is more substantial. The temperature scaling has pushed the neural network to have postive BSS, meaning the calibration is better than a model that outputs the prevalence of the events. However we should note that both of the post hoc calibration techniques have lowered the AUC-PRC, meaning the predictive power is compromised for the calibration. In our model we can observe high predictive power as well as calibration. We postulate that the decoupling training indeed separates the bias from imbalanced data to the classifier while the feature extractor maintains the power of absorbing all information. The label distribution-aware loss, acting similarly to the Temperature scaling (where the scaling factor is inherently tuned during training with our modified softmax formulation), is calibrating the balanced classifier without sacrifice of predictive power. To this end, without using the extra calibration dataset is not an issue anymore. We have tried to test the Temperature scaling on our model but we did not observe improvement on calibration but the AUC-ROC and AUC-PRC are slightly compromised as well.

\begin{figure}[t!]
\centering
\includegraphics[width=0.55\textwidth]{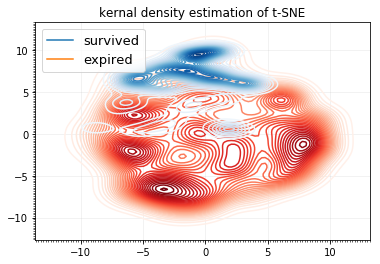}
\caption{Density plot of survived/expired patients}
\label{fig:density}
\end{figure}

Furthermore, as a proof-of-concept, we are particularly interested if our assumption holds, i.e. the patients who survived would be similar to each other where the patients who expire would be more dissimilar. We have extracted the embedded vector from our model whose backbone is based on \cite{deasy2019impact}, carrying 256-dimension last hidden layer (before classifier) on the mortality prediction task in MIMIC-III. And then we apply t-SNE \cite{linderman2019fast} which is a visualization algorithm to embed the data into 2 dimensions. We plot the embedding along with their labels to show the density differences, shown in Figure \ref{fig:density}. As we can see, the survived patients account for a small and condensed space where the patients who expired would form different peaks, indicating different local clusters (e.g. diseases/phenotypes). This can prove the assumption of the density discrepancy as training information can be truly captured.

\begin{table*}[t!]
\small
\centering
\caption{Ablation study for TOPCAT dataset}
\label{tab:TOCAT_ablation}
\begin{tabular}{|l|l|l|l|l|}
\hline
Task & Methods  & AUC-ROC & AUC-PRC & BSS  \\ \hline
\multirow{5}{*}{Mortality} 
& MLP     & 0.736 $\pm$ 0.004  &  0.523 $\pm$ 0.002   &    -0.067  $\pm$ 0.001      \\
& MLP-TrainableCost     & 0.770 $\pm$ 0.003  &  0.541 $\pm$ 0.002   &    -0.188  $\pm$ 0.001      \\
& MLP-decoupling     & 0.778 $\pm$ 0.002  &  0.569 $\pm$ 0.005   &   -0.480 $\pm$ 0.004       \\
& MLP-FL     & 0.782 $\pm$ 0.004  &  0.541 $\pm$ 0.003   &   -0.080 $\pm$ 0.004       \\
& MLP-DAH     & 0.779 $\pm$ 0.001  &  0.549 $\pm$ 0.005   &   0.111 $\pm$ 0.004      \\
\cline{2-5}
& MLP-Ours  & \textbf{0.798 $\pm$ 0.002}  &  \textbf{0.589 $\pm$ 0.001}   &      \textbf{0.178 $\pm$ 0.002}     \\
\hline
\end{tabular}
\end{table*}

\begin{table*}[t!]
\small
\centering
\caption{Ablation study for MIMIC-III dataset}
\label{tab:MIMIC_ablation}
\begin{tabular}{|l|l|l|l|l|}
\hline
Task & Methods & AUC-ROC & AUC-PRC & BSS  \\ \hline
\multirow{5}{*}{Mortality} 

& GRU   & 0.871 $\pm$ 0.004  &  0.514 $\pm$ 0.003  &  -1.116      $\pm$ 0.005   \\
& GRU-TrainableCost   & 0.879 $\pm$ 0.001  &  0.520 $\pm$ 0.002  &  -1.108      $\pm$ 0.005   \\
& GRU-decoupling   & \textbf{0.892 $\pm$ 0.003} &  0.577 $\pm$ 0.002  &  -0.909   $\pm$ 0.002     \\
& GRU-FL    & 0.875 $\pm$ 0.008 &  0.523$\pm$ 0.007 &  -0.112 $\pm$ 0.007       \\
& GRU-DAH     & 0.876 $\pm$ 0.003 &  0.534 $\pm$ 0.005 &  0.078   $\pm$ 0.003      \\
\cline{2-5}
& GRU-Ours    & \textbf{0.892$\pm$ 0.001}  & \textbf{0.586$\pm$ 0.004}   &   \textbf{0.240$\pm$ 0.004}      \\
\hline 
\end{tabular}
\end{table*}

\subsubsection{Ablation Study}
We have a few components in our model such as decoupling training and density aware loss function. We are interested to know what makes the model improve and how can we dissect the model to demonstrate. We aim to study how does the decoupling help the prediction, and specifically what has the model learned. Also by comparing density aware loss with vanilla version (traditional cross-entropy loss) as well as another advanced version of loss function (focal loss \cite{lin2017focal}), we conduct a thorough comparison between them. In MIMIC, we have an existing strong backbone that we can apply our techniques on \cite{deasy2019impact}, which is based on a GRU model. In the TOPCAT dataset, to construct a strong backbone, we make use of the same MLP architecture as in R-MLP \cite{babar2016mlp} with a residual skip connection block \cite{he2016deep} that can be further decoupled or trained with different loss functions. We listed our ablation study in Table \ref{tab:TOCAT_ablation} and \ref{tab:MIMIC_ablation} in these two datasets both for mortality prediction.

\begin{figure}[b]
\centering
\begin{minipage}{.7\textwidth}
\centering
  \vspace*{6mm}
  \includegraphics[width=\textwidth]{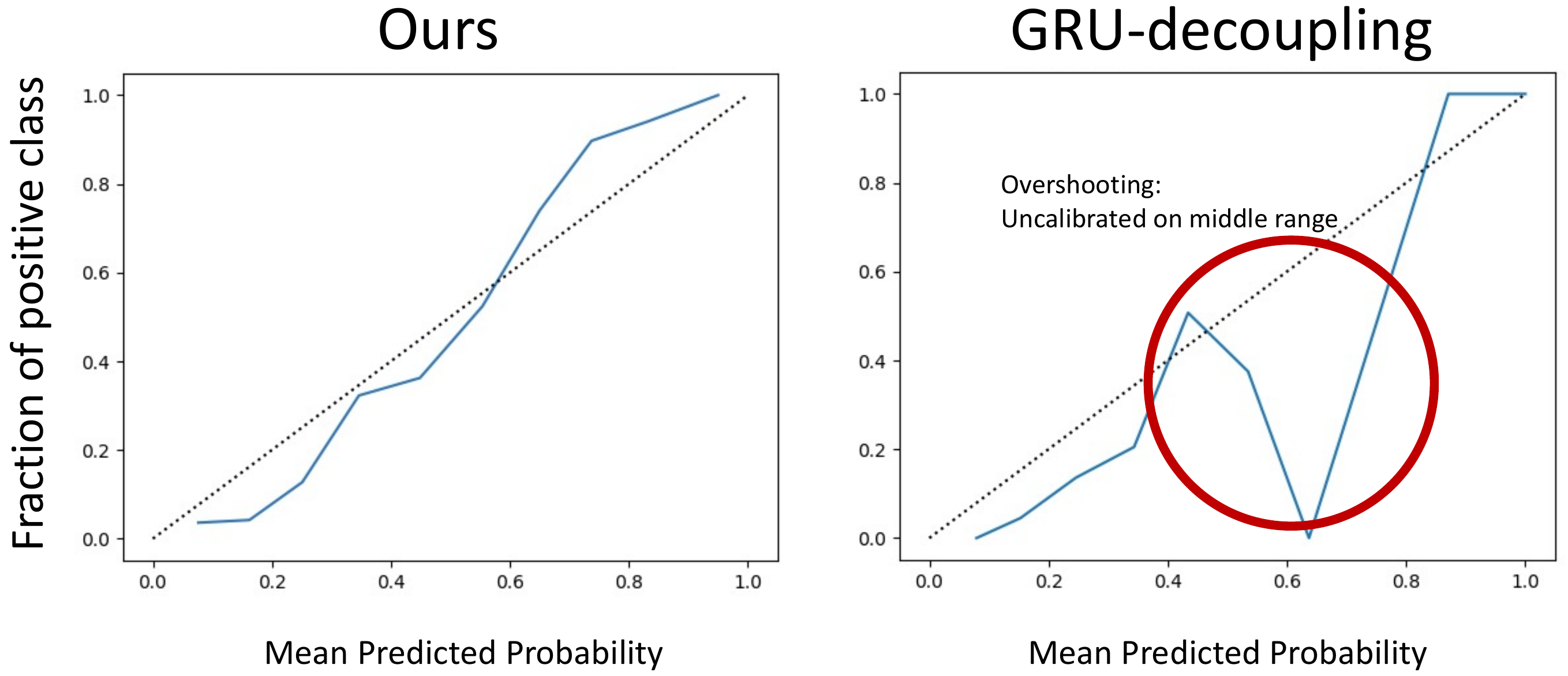}\vspace*{-3mm}
  \caption{Calibration plot comparison}
  \label{fig:calibration}
\end{minipage}
\end{figure}

First, for the TOPCAT dataset, we can see that when fully applying our framework on the backbone, the model would outperform all other variants in Table \ref{tab:TOCAT_ablation}. Another finding is that the decoupling training is improving the AUC-PRC in a larger margin than others, suggesting that this way of training can largely avoid the imbalance issue by through a more distribution-aware metric. However the shortcoming of decoupling alone is that it is bad at calibration, where it is among the worst BSS metric in the methods. Second, when applying density aware loss alone, we can see the model can be better calibrated (i.e. positive BSS), which is usually an important aspect of a medical model \cite{angraal2020machine} because the output probability can be evaluated as the risk score for further ranking. For the MIMIC-III dataset in Table \ref{tab:MIMIC_ablation}, we can see that the decoupling itself can improve significantly and this method alone can give good AUC-ROC (tied as best). We are then interested to know how does this single trick compare to the full framework on improving BSS in terms of calibration. The comparison is in Fig \ref{fig:calibration} where we can see our model's calibration is closer to diagonal, rendering a more natural  `S' shape \cite{fellowship2008b}, where the baseline GRU-decoupling has poor calibrated range when the output probability is in mid/high range (which is the label of high-risk patients). This is showing the model is overshooting for this range of probability, likely due to an density discrepancy, because the model would assign overconfident probability to the patients in higher risk, requiring a density aware training. The over-confident prediction is prevalent in modern-day neural networks, where the mean predicted probability is higher than the fraction of positive class in a certain bin as noted in \cite{guo2017calibration}. However, when equipped with the full framework, the performance of our model can increase significantly, especially on Brier Skill Score for calibration and rendering the plot to be closer to the diagonal (perfect calibration).

\vspace{-1mm}
\subsubsection{Parameter study}
\begin{figure}[ht!]
\centering
\includegraphics[width=0.4\textwidth]{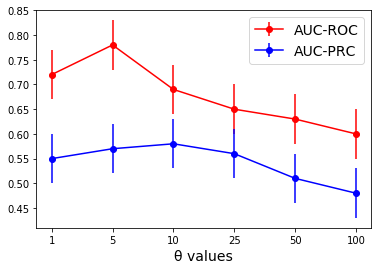}\vspace{-4mm}
\caption{Tuning of $\theta$ for AUC-ROC and AUC-PRC}
\label{fig:cost}
\end{figure}

Since we have incorporated a trainable cost matrix, and we are interested in how does the parameter $\theta$ in Eq. \ref{trainable_cost} change the performance in the model. We have search on a space of $\{1, 5, 10, 25, 50, 100\}$ for $\theta$, following \cite{roychoudhury2017cost}. On the TOPCAT dataset for mortality prediction we conduct the experiments and show it in Figure \ref{fig:cost}. We can see that AUC-ROC peaks at $\theta=5$ while AUC-PRC can be $\theta=10$. However, given the confidence interval's overlap, the significance for choosing $\theta=10$ over $\theta=5$ for AUC-PRC can be statistically minimal, therefore $\theta=5$ is chosen.

\vspace{-6mm}
\section{Conclusion}
We proposed a framework to treat class imbalance, which is prevalent in medical datasets. The introduced framework not only addresses imbalanced class densities but also makes use of the density discrepancy to train a model. The decoupled training method alleviated bias caused by the majority class, by ensuring faithful  representation of the minority class. Further, we used a density-aware loss to personalize training of each class, specifically: learning that lower-risk patients arrive at low risk by calculation of the similar factors, forming a dense cluster in the data space, but high-risk patients are dissimilar, driving them to different regions of the data space. We demonstrated that our model, trained with this decoupling framework along with density-aware loss and learnable cost matrix, outperformed baseline approaches when applied to risk prediction in medical datasets. Furthermore, through experiments we find that traditional models were poorly calibrated, calling for more comprehensive evaluation, especially geared towards imbalance issues. Our framework overall has shown to be better at prediction as well as calibration, which can be of great use in the medical domain. 
\acks{This project is in part supported by the Defense Advanced Research Projects Agency under grant FA8750-18-2-0027.}

\small
\bibliography{bib}

\appendix
\section*{Appendix A.}

\vspace{3mm}
The table of used variables for TOPCAT dataset along with their definitions

\begin{longtable}{| p{.3\textwidth} | p{.70\textwidth} |} 

\hline
Variable Names	&	Definition	\\
\hline
age\textunderscore entry.x	&	Age entering the study	\\
GENDER.x	&	Gender of the subject	\\
RACE\textunderscore WHITE	&	White or Caucasian	\\
RACE\textunderscore BLACK	&	Race: Black	\\
RACE\textunderscore ASIAN	&	Race: Asian	\\
RACE\textunderscore OTHER	&	Race: Other	\\
ETHNICITY	&	Subject of Hispanic, Latino, or Spanish origin?	\\
DYSP\textunderscore CUR	&	Dyspnea: Present at screening?	\\
DYSP\textunderscore YR	&	Dyspnea: experienced in past year?	\\
ORT\textunderscore CUR	&	Orthopnea: Present at screening?	\\
ORT\textunderscore YR	&	Orthopnea: experienced in past year?	\\
DOE\textunderscore CUR	&	Dyspnea on exertion: Present at screening?	\\
DOE\textunderscore YR	&	Dyspnea on exertion: experienced in past year?	\\
RALES\textunderscore CUR	&	Rales present at screening?	\\
RALES\textunderscore YR	&	Rales: experienced in past year?	\\
JVP\textunderscore CUR	&	JVP: Present at screening?	\\
JVP\textunderscore YR	&	JVP: experienced in past year?	\\
EDEMA\textunderscore CUR	&	Edema: Present at screening?	\\
EDEMA\textunderscore YR	&	Edema: experienced in past year?	\\
EF	&	Ejection Fraction	\\
CHF\textunderscore HOSP	&	Previous hospitalization for CHF	\\
chfdc\textunderscore dt3	&	Time Between randomization and Hospitalization for Cardiac Heart Failure (years)	\\
MI	&	Previous myocardial infarction	\\
STROKE	&	Previous Stroke	\\
CABG	&	Previous Coronary artery bypass graft surgery	\\
PCI	&	Previous Percutaneous Coronary Revascularization	\\
ANGINA	&	Angina Pectoris	\\
COPD	&	Chronic Obstructive Pulmonary Disease	\\
ASTHMA	&	Asthma	\\
HTN	&	Hypertension	\\
PAD	&	Peripheral Arterial Disease	\\
DYSLIPID	&	Dyslipidemia	\\
ICD	&	Implanted cardioverter defibrillator	\\
PACEMAKER	&	Pacemaker implanted	\\
AFIB	&	Atrial fibrillation	\\
DM	&	Diabetes Mellitus	\\
treat\textunderscore sp\textunderscore cat	&	Treat for diabetes mellitus: other: specify (categorical variable)	\\
SMOKE\textunderscore EVER	&	Has subject ever been a smoker	\\
QUIT\textunderscore YRS	&	How many years since quitting	\\
alcohol4\textunderscore cat	&	How many Drinks do you consume per week (0/1-5/5-10/11+)	\\
HEAVY\textunderscore WK	&	Exercise: Heavy	\\
MED\textunderscore WK	&	Exercise: Medium	\\
LIGHT\textunderscore WK	&	Exercise: Light	\\
LIGHT\textunderscore MIN	&	Exercise: Light: Minutes	\\
mets per week	&	Activity Level (mets per week)	\\
cooking\textunderscore salt\textunderscore score	&	Cooking Salt Score	\\
nyha\textunderscore class\textunderscore cat	&	NYHA class 3\&4 vs 1\&2	\\
HR.x	&	Heart rate	\\
SBP	&	Systolic blood pressure	\\
DBP	&	Diastolic blood pressure	\\
gfr	&	Glomerular Filtration Rate	\\
NA\textunderscore mmolL	&	Sodium: Result (mmol/L)	\\
K\textunderscore mmolL	&	Potassium: Result (mmol/L)	\\
CL\textunderscore mmolL	&	Chloride: Result (mmol/L)	\\
CO2\textunderscore mmolL	&	CO2: Result (mmol/L)	\\
BUN\textunderscore mgdL	&	Blood Urea Nitrogen: Result (mg/dL)	\\
GLUCOSE\textunderscore mgdL	&	Glucose: Result (mg/dL)	\\
GLUCOSE\textunderscore INDICATOR	&	Whether the glucose measured was fasting or random	\\
WBC\textunderscore k/$\mu$L	&	WBC count: Result (k/uL)	\\
HB\textunderscore gdL	&	Hemoglobin: Result (g/dL)	\\
PLT\textunderscore k/$\mu$L	&	Platelet Count: Result (k/uL)	\\
ALT\textunderscore UL	&	Alanine Aminotransferase: Results (U/L)	\\
ALP\textunderscore UL	&	Alkaline Phosphatase: Results (U/L)	\\
AST\textunderscore UL	&	Aspartate Aminotransferase: Results (U/L)	\\
TBILI\textunderscore mgdL	&	Total Bilirubin: Results (mg/dL)	\\
ALB\textunderscore gdL	&	Albumin: Results (g/dL)	\\
urine\textunderscore val\textunderscore mgg	&	Urine Microalbumin/Creatinine Ratio: Result (mg/g)	\\
QRS\textunderscore DUR	&	QRS Duration	\\
ECG\textunderscore AFIB	&	Atrial fibrillation/Flutter	\\
ECG\textunderscore BBB2	&	Bundle Branch Block - Yes/No indicator	\\
ECG\textunderscore VPR	&	Ventricular paced rhythm	\\
ECG\textunderscore Q	&	Pathological Q waves	\\
ECG\textunderscore LVH	&	Left ventricular hypertrophy	\\
drug.x	&	Treatment Group (Spironolactone or Placebo)	\\
BMI	&	Body Mass Index	\\
cigpacksperday	&	Number of cigarettes per day	\\
phys\textunderscore limit\textunderscore score	&	KCCQ: Physical Limitation score	\\
symp\textunderscore stab\textunderscore score	&	KCCQ: Symptom Stability score	\\
symp\textunderscore freq\textunderscore score	&	KCCQ: Symptom Frequency score	\\
symp\textunderscore bur\textunderscore score	&	KCCQ: Symptom Burden score	\\
tot\textunderscore symp\textunderscore score	&	KCCQ: Total Symptom score	\\
self\textunderscore eff\textunderscore score	&	KCCQ: Self-Efficacy score	\\
qol\textunderscore score	&	KCCQ: Quality of Life score	\\
soc\textunderscore limit\textunderscore score	&	KCCQ: Social Limitation score	\\
overall\textunderscore sum\textunderscore score	&	KCCQ: Overall Summary score	\\
clin\textunderscore sum\textunderscore score	&	KCCQ: Clinical Summary score	\\
\hline
\caption{List of Candidate Variables used for Predicting Mortality }
\label{tab:TOPCAT_variables}
\end{longtable}

\begin{table*}[h] 
\caption{Phenotype labels for MIMIC-III dataset}
\begin{center}
\begin{normalsize}
\begin{tabular}{c|c}
\hline
\hline
Phenotype & type  \\
\hline
{ \footnotesize{Acute and   unspecified renal failure}}     & { \footnotesize{acute}}   \\ 
{ \footnotesize{Essential hypertension}}                     & { \footnotesize{chronic}} \\
{ \footnotesize{Acute cerebrovascular disease}}             & { \footnotesize{acute}}    \\ 
{ \footnotesize{Fluid and electrolyte disorders}}            & { \footnotesize{acute}}    \\
{ \footnotesize{Acute myocardial infarction}}               & { \footnotesize{acute}}    \\ 
{ \footnotesize{Gastrointestinal hemorrhage}}                & { \footnotesize{acute}}    \\
{ \footnotesize{Respiratory failure; insufficiency; arrest}}& { \footnotesize{acute}}    \\ 
{ \footnotesize{Hypertension with complications}}            & { \footnotesize{chronic}}  \\
{ \footnotesize{Chronic kidney disease}}                    & { \footnotesize{chronic}} \\
{ \footnotesize{Other liver diseases}}                       & { \footnotesize{mixed}}    \\
{ \footnotesize{Chronic obstructive pulmonary disease}}     & { \footnotesize{chronic}}  \\ 
{ \footnotesize{Other lower respiratory disease}}            & { \footnotesize{acute}}    \\
{ \footnotesize{Complications of surgical/medical care}}    & { \footnotesize{acute}}   \\ 
{ \footnotesize{Other upper respiratory disease}}            & { \footnotesize{acute}}     \\
{ \footnotesize{Pleurisy; pneumothorax; pulmonary collapse}}& { \footnotesize{acute}}    \\ 
{ \footnotesize{Conduction disorders}}                      & { \footnotesize{mixed}}    \\
{ \footnotesize{Congestive heart failure; nonhypertensive}} & { \footnotesize{mixed}}    \\ 
{ \footnotesize{Pneumonia}}                                  & { \footnotesize{acute}}    \\
{ \footnotesize{Coronary atherosclerosis and related}}      & { \footnotesize{chronic}}  \\ 
{ \footnotesize{Cardiac dysrhythmias}}                      & { \footnotesize{mixed}}    \\
{ \footnotesize{Diabetes mellitus with complications}}      & { \footnotesize{mixed}}    \\
{ \footnotesize{Diabetes mellitus without complication}}    & { \footnotesize{chronic}} \\
{ \footnotesize{Disorders of lipid metabolism}}             & { \footnotesize{chronic}} \\
{ \footnotesize{Septicemia (except in labor)}}               & { \footnotesize{acute}}   \\
{ \footnotesize{Shock}}                                      & { \footnotesize{acute}}       \\
\hline
\label{MIMIC_pheno}
\end{tabular}
\end{normalsize}
\end{center}
\end{table*}

\end{document}